\documentclass[letterpaper, 10 pt, conference]{ieeeconf}
\IEEEoverridecommandlockouts
\usepackage{cite}
\usepackage{amsmath,amssymb,amsfonts}
\usepackage{algorithmic}
\usepackage{graphicx}
\usepackage{adjustbox}
\usepackage{textcomp}
\usepackage{xcolor}
\usepackage{caption,subcaption}
\usepackage{makecell}
\usepackage{soul}

\def\BibTeX{{\rm B\kern-.05em{\sc i\kern-.025em b}\kern-.08em
    T\kern-.1667em\lower.7ex\hbox{E}\kern-.125emX}}
\begin{document}

\title{\LARGE \bf Trajectory Prediction in Autonomous Driving with a Lane Heading Auxiliary Loss}

\author{Ross Greer$^{1}$, Nachiket Deo$^{1}$, and Mohan Trivedi$^{2}$
\thanks{$^{1}$Ross Greer and Nachiket Deo are graduate students in the Laboratory for Intelligent \& Safe Vehicles (LISA) at University of California, San Diego. {\tt\small regreer@eng.ucsd.edu, ndeo@eng.ucsd.edu}}%
\thanks{$^{2}$Mohan Trivedi is a Distinguished Professor of Electrical and Computer Engineering at University of California, San Diego, and Director of LISA.
        {\tt\small mtrivedi@eng.ucsd.edu}}%
}

\maketitle

\begin{abstract}
Predicting a vehicle's trajectory is an essential ability for autonomous vehicles navigating through complex urban traffic scenes. Bird's-eye-view roadmap information provides valuable information for making trajectory predictions, and while state-of-the-art models extract this information via image convolution, auxiliary loss functions can augment patterns inferred from deep learning by further encoding common knowledge of social and legal driving behaviors. Since human driving behavior is inherently multimodal, models which allow for multimodal output tend to outperform single-prediction models on standard metrics. We propose a loss function which enhances such models by enforcing expected driving rules on all predicted modes. Our contribution to trajectory prediction is twofold; we propose a new metric which addresses failure cases of the off-road rate metric by penalizing trajectories that oppose the ascribed heading (flow direction) of a driving lane, and we show this metric to be differentiable and therefore suitable as an auxiliary loss function. We then use this auxiliary loss to extend the the standard multiple trajectory prediction (MTP) and MultiPath models, achieving improved results on the nuScenes prediction benchmark by predicting trajectories which better conform to the lane-following rules of the road. 
\end{abstract}


\section{Introduction}
\label{sec:intro}

\begin{figure}[t]
\centering
\includegraphics[width=\columnwidth]{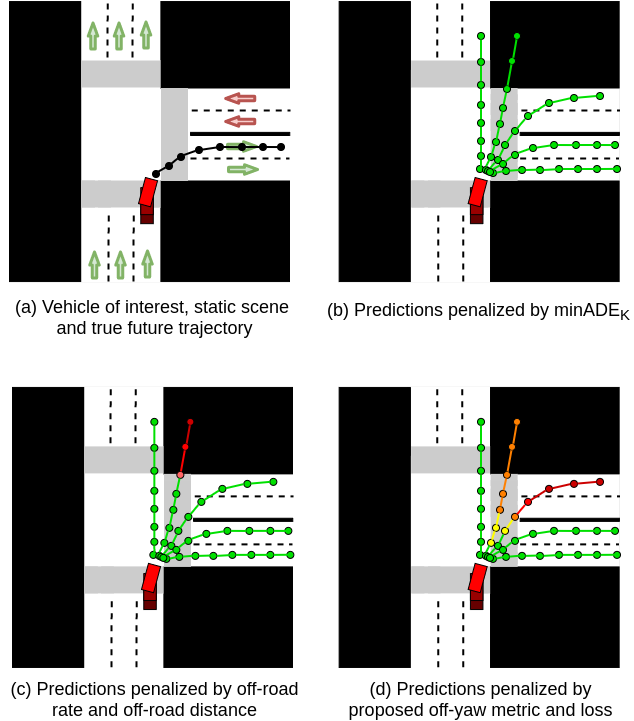}
\caption{\textbf{Motivating example:} A vehicle of interest approaching an intersection (top-left). The commonly used minADE$_K$ metric fails to penalize a diverse set of poor trajectories (top-right). The off-road rate and off-road distance metrics partially address this (bottom-left), but fail to penalize trajectories that violate lane direction. Our proposed off-yaw metric and corresponding YawLoss seek to address this (bottom-right). Severity of imposed penalty is illustrated by color, with green minimal and red maximal.}
\label{fig:concept}
\end{figure}

To safely navigate complex city traffic, autonomous vehicles need the ability to predict the future trajectories of surrounding vehicles. There is inherent uncertainty in predicting future trajectories, making it a challenging task. In particular, the distribution of future trajectories is multimodal. At a given instant in a traffic scene, a driver could have one of several plausible goals, with multiple paths to each goal. 

Recent work has addressed multimodality in trajectory prediction by learning models that output multiple trajectories conditioned on the past motion of agents and the static scene around them. Common approaches include learning mixture models \cite{deo2018multi, deo2018convolutional, cui2019multimodal, IEEEexample:multipath, ridel2020scene, messaoud2020multi, casas2020importance, liang2020learning}, sampling latent variable models \cite{gupta2018social, sadeghian2019sophie, amirian2019social, kosaraju2019social, zhao2019multi, wang2020improving, lee2017desire, ivanovic2019trajectron, salzmann2020trajectron++, casas2020implicit, rhinehart2018r2p2, rhinehart2019precog, bhattacharyya2019conditional, bhattacharyya2020haar}, or sampling stochastic policies trained using inverse reinforcement learning \cite{ziebart2009planning, kitani2012activity, zhang2018integrating, deo2020trajectory}. However, defining appropriate evaluation metrics for models that output multiple trajectories still remains an open challenge.

The most commonly used evaluation metric for multimodal trajectory prediction is the minimum average displacement error over $K$ trajectories (minADE$_K$). This has the advantage of not penalizing diverse, but plausible trajectories output by models. A limitation of minADE$_K$ is that it fails to penalize models that output a diverse set of trajectories of poor quality (Fig \ref{fig:concept}b). This has been addressed in prior work by additionally reporting sample quality metrics. Of particular interest are the off-road rate and off-road distance metrics \cite{niedoba2019improving, boulton2020motion} which penalize predictions that fall outside the drivable area in a scene, visualized in Fig \ref{fig:concept}c. However, there's more structure to vehicle motion: vehicles typically follow the direction ascribed to lanes. A naive formulation of the off-road rate or off-road distance metrics fails to penalize trajectories wrongly predicted in the direction of oncoming traffic.

In this work, we define a new metric for sample quality of predicted trajectories termed the \textit{off-yaw rate}. The off-yaw rate measures the adherence of predicted trajectories to lane direction, and penalizes predictions that violate lane direction (Fig \ref{fig:concept}d).
Moreover, we show that the off-yaw rate can be used as a differentiable loss function termed \textit{YawLoss}, which can serve as an auxiliary training loss for multimodal trajectory prediction models. Our formulation of the YawLoss can be applied for training both mixture models as well as latent variable models for trajectory prediction, and leads to predicted trajectories that better conform to the lane direction, while also achieving lower minADE$_K$ values. We report results on the publicly available NuScenes prediction benchmark by incorporating the YawLoss for training two vehicle trajectory prediction models that represent the state of the art, namely MTP proposed by Cui \textit{et al.} \cite{cui2019multimodal}  and Multipath proposed by Chai \textit{et al.}\cite{IEEEexample:multipath}.

\section{Related Research}

\subsection{Multimodal trajectory prediction}
A large body of recent literature has addressed the problem of human and vehicle trajectory prediction. For comprehensive surveys we refer the reader to \cite{rudenko2020human, ridel2018literature}. Here, we discuss models that output multimodal predictions. A common approach for multimodal trajectory prediction is to learn mixture models. Each mixture component represents a mode of the trajectory distribution. Models typically output mean trajectories for each mode and standard deviations, along with a categorical probability distribution over modes. Early work associated modes of the trajectory distribution with pre-defined maneuvers or intents \cite{deo2018multi, deo2018convolutional}. The need for pre-defined maneuvers was alleviated by the multiple trajectory prediction (MTP) loss proposed by Cui \textit{et al.} \cite{cui2019multimodal}. The MTP loss has a cross-entropy component for learning the categorical probability distribution over modes, and a regression component that only penalizes the mode that is closest to the ground truth. 

This formulation has since been used by subsequent works \cite{ridel2020scene, messaoud2020multi, liang2020learning, casas2020importance}. More recently Chai \textit{et al.} \cite{IEEEexample:multipath} extended this idea to learn deviations from anchor trajectories as modes of the trajectory distribution, rather than mean trajectories themselves, and Phan-Minh \textit{et al.} \cite{phan2020covernet} proposed to discard regression outputs altogether while just assigning probabilities to a discrete trajectory set. Another common approach for multimodal trajectory forecasting is learning latent variable models. Conditioned on input context such as past trajectories and static scene, latent variable models map samples from a simple latent distribution to trajectory samples. Prior works have used generative adversarial networks \cite{gupta2018social, sadeghian2019sophie, amirian2019social, wang2020improving, kosaraju2019social, zhao2019multi}, conditional variational autoencoders \cite{lee2017desire, ivanovic2019trajectron, salzmann2020trajectron++, casas2020implicit}, and more recently normalizing flow based models \cite{rhinehart2018r2p2, rhinehart2019precog, bhattacharyya2019conditional, bhattacharyya2020haar}. Finally, some approaches output multimodal predictions by sampling stochastic policies learned using inverse reinforcement learning \cite{ziebart2009planning, kitani2012activity, zhang2018integrating, deo2020trajectory}. 

While our proposed off-yaw metric and YawLoss can be used in conjunction with any approach that involves regression outputs, here we report results using the MTP and Multipath models as baselines. Both models aim to predict the most likely trajectory of a vehicle from a set of trajectories output by a neural network and their respective probabilities. In the MTP network, a rasterized map containing an overhead view of the surrounding roadyway and vehicles is passed through a CNN backbone, then flattened and concatenated with the ego vehicle's state vector (velocity, acceleration, and heading rate change). This combined vector is then passed through a series of fully-connected layers, ending with an output of $M$ modes comprised of $2H+1$ values each, representing the $H$ $(x, y)$-values per trajectory plus an associated probability. Similarly, the Multipath model takes the same rasterized map as input, but utilizes a crop around the ego vehicle in between convolutional layers to better feed relevant mid-level features forward in the network. However, the Multipath approach makes use of pre-computed anchors, taken to be the $K$-mean clusters (or alternative cluster methodology) of the training set trajectories. The network will output $M$ modes comprised of $5H+1$ values each, representing the offset from the anchor in the x and y directions, the three parameters used to define the covariance matrix for the prediction, and the associated mode probability. 

\subsection{Sample quality metrics and auxiliary loss functions for trajectory prediction}
As described in section \ref{sec:intro}, the commonly used minADE$_K$ metric for trajectory prediction is a good measure for sample diversity, but can be a poor measure of sample quality or precision. There is inherent tension between sample diversity and sample quality or precision \cite{rhinehart2018r2p2}. Several works have thus employed metrics in addition to minADE$_K$ for measuring sample quality of trajectories. Rhinehart \textit{et al.} \cite{rhinehart2018r2p2} define a symmetric KL divergence metric with a component that measures sample diversity, and a component that measures sample precision and also use both metrics as loss functions for training. Some works \cite{casas2020spagnn, casas2020implicit, rhinehart2019precog} report collision rates for trajectories predicted for multiple actors in the scene, penalizing falsely predicted collisions. Cui \textit{et al.} \cite{cui2020deep} report kinematic feasibility of predicted vehicle trajectories. Casas \textit{et al.} \cite{casas2020importance} report lane infractions via traffic light or lane divider violations in predicted trajectories, as well as performance metrics for a downstream planner relying on these predictions. They also use prior knowledge of reachable lanes and the route of the autonomous vehicle to define a reward function for training the trajectory prediction model via the REINFORCE algorithm. Finally, closely related to our work, a large number of approaches use the off-road rate and off-road distance metrics \cite{niedoba2019improving, deo2020trajectory, boulton2020motion, chang2019argoverse, ridel2020scene, messaoud2020multi} for evaluating predicted trajectories. These metrics compute the proportion of predicted points that lie outside the drivable region of the road and the nearest distance of predicted points to the drivable region respectively. Niedoba \textit{et al.} \cite{niedoba2019improving}, Boulton \textit{et al.} \cite{boulton2020motion} and Messaoud \textit{et al.} \cite{messaoud2020multi} also use the off-road rate as a loss function for training trajectory prediction model. Our off-yaw rate and YawLoss improve upon the off-road rate by explicitly reasoning about the direction of motion of lanes and penalizing predicted trajectories that violate it.

\section{Off-Yaw Rate as a Metric}

\subsection{Off-Yaw Rate}
By accepted legal and social convention, when driving in a lane, the vehicle must move in the direction of the lane heading as to not interfere with other traffic. The off-yaw rate is a measure of a trajectory's ability to orient in the direction of the nearest lane.  

Define a vehicle's initial position on trajectory $\tau$ as $(x_0^{\tau}, y_0^{\tau}) = (0, 0)$, and its initial orientation in the local frame as $ \theta = 0 $ aligned with the standard y-axis. Given a trajectory of points $\tau = \{(x^{\tau}_0, y^{\tau}_0), (x^{\tau}_1, y^{\tau}_1), ..., (x^{\tau}_n, y^{\tau}_n)\}$, where points 1 through $n$ correspond to predicted future points, we can estimate the vehicle heading relative to its initial orientation with the following procedure. First, we assume the trajectory sample rate relative to map scale is sufficiently high that we can accept a straight-line approximation between consecutive points. Let $(\hat{x}^{\tau}_{i}, \hat{y}^{\tau}_{i})$ be the midpoint of two consecutive points $(x^{\tau}_i, y^{\tau}_i), (x^{\tau}_{i+1},y^{\tau}_{i+1})$, defined by the function:

\begin{equation}
    (\hat{x},\hat{y})(x_1, y_1, x_2, y_2) = (\frac{x_1+x_2}{2}, \frac{y_1+y_2}{2})
\end{equation}

The angle between the same two consecutive trajectory points surrounding $(\hat{x}^{\tau}_{i}, \hat{y}^{\tau}_{i})$ is found using 
\begin{equation}
    \theta(x_1, y_1, x_2, y_2) = \arctan{(\frac{x_{2} - x_1}{y_2 - y_1})}.
\end{equation}

\begin{figure}
    \centering
    \includegraphics[width=40mm]{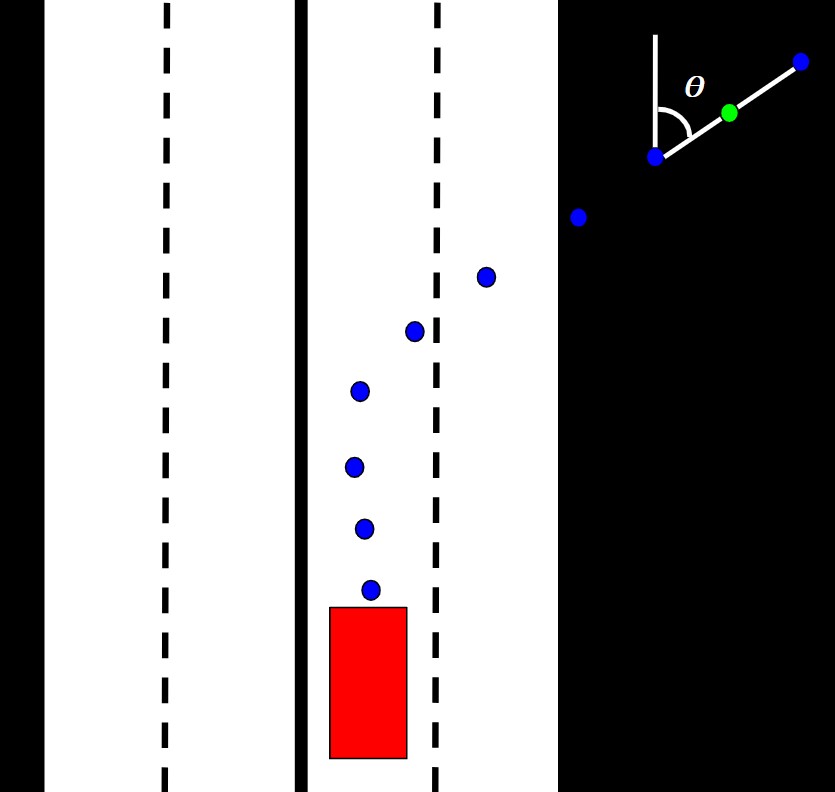}
    \caption{The predicted trajectory of the ego vehicle (red) is shown in blue. The green circle represents a midpoint $i$ between two points of the trajectory. The angle $\theta_i$, in the local frame, is assigned to midpoint $i$.}
    \label{fig:my_label2}
\end{figure}

This angle $\theta(x^{\tau}_i, y^{\tau}_i, x^{\tau}_{i+1},y^{\tau}_{i+1})$ is then paired with the midpoint $(\hat{x}^{\tau}_{i}, \hat{y}^{\tau}_{i})$, illustrated in Fig. \ref{fig:my_label2}. From a series of $n$ estimated trajectory points, we create a series of $n$ midpoints and associated headings relative to the initial orientation, which can be converted directly from the local frame to the global frame using the ego vehicle's rotation matrix. We refer to the $i$-th heading of a trajectory in the local frame as $\theta_{\tau,i}$, and the same heading in the global frame as $\theta^{G}_{\tau,i}$.

The angular difference between a trajectory midpoint heading in the global frame, $\theta$ and the heading of the nearest lane, $\theta_{NL}(x,y)$ can be calculated as follows:

\begin{equation}
    \delta(x,y,\theta) = min(\theta - \theta_{NL}(x,y), \theta_{NL}(x,y) - \theta).
\end{equation}

A successful measure of off-yaw driving should increase for any portion of the trajectory $\tau$ which deviates from the lane orientation. Further, greater angular differences should be assigned greater values than smaller angular differences. The off-yaw measure of an $n$-point trajectory is:

\begin{equation}
    Y(\tau) = \sum_{i=1}^{n}{\delta(\hat{x}^{\tau}_i, \hat{y}^{\tau}_i, \theta_{\tau,i}^G)}.
\end{equation}

Extending over all $m$ predicted modes, we reach the per-sample average off-yaw expression: 

\begin{equation}
    Y = \sum_{\tau=1}^{m}{Y(\tau)}
\end{equation}

\subsection{Lane Change Approximations}

There is a small margin of expected angular error, $ \epsilon $, for minor adjustments to the vehicle heading in order to stay within the lane. In addition to lane-correcting error $\epsilon$, a second exception to the assumption of lane-aligned driving occurs when a driver changes lanes, during which their vehicle may orient at an angle no more than (and typically much less than) $90^{\circ}$ to perform the lane change maneuver, with a $90^{\circ}$ lane change occurring only when traffic is at a stop. Typical lane changes occur at angles relative to the flow of traffic and vehicle dynamics such as turning radius and velocity.
Since a trajectory should not be considered off-yaw during a legal lane change, nor during small-angle lane corrections, we therefore constrain the measure function to only penalize angular differences which exceed a threshold, $\alpha$. The modified angular difference, $\hat{\delta}_{i}$, has the following formula: 

\begin{equation}
    \delta^{\alpha}(x,y,\theta)  = 
    \begin{cases} 
      0 & \delta(x,y,\theta) \leq \alpha \\
      \delta(x,y,\theta) & \delta(x,y,\theta) > \alpha \\
   \end{cases}
\end{equation}

For our experiments, we selected a threshold of $45^{\circ}$. 

\subsection{Off-Yaw in Intersections}

When a vehicle passes through an intersection, the vehicle must cross over lanes which flow in discordant directions (look no further than the existence of stoplights as proof). At these moments, the nearest lane point to the vehicle may belong to a lane which flows in opposite direction, even though it is perfectly reasonable for the vehicle to be in this position. To account for these situations, the measure should not penalize deviation from the heading of the closest lane for midpoints which lie in an intersection. Thus, the measure is modified to drop values which occur in an intersection:  

\begin{equation}
    Y^{\alpha}(\tau) = \frac{1}{n}\sum_{i=1}^{n}{I(x^{\tau}_i,y^{\tau}_i)\delta^{\alpha}(x^{\tau}_i, y^{\tau}_i)},
\end{equation}

where

\begin{equation} \label{eq:indic}
    I(x^{\tau}_i, y^{\tau}_i) = 
    \begin{cases}
    0 & (x^{\tau}_{i}, y^{\tau}_{i}) \text{ in intersection} \\
    1 & \text{otherwise}
    \end{cases}
\end{equation}.

Summing the values computed for all $m$ predicted modes, we reach the modified per-sample off-yaw measure expression: 

\begin{equation} \label{eq:offyaw}
    \bar{Y}^{\alpha}(T) = \sum_{\tau=1}^{m}{Y^{\alpha}(\tau)}
\end{equation}

The off-yaw rate for a set of samples and their predicted trajectory sets is the average fraction of trajectories which contain off-yaw events, defined in the following equation:

\begin{equation}
    R_{\text{off-yaw}} = \frac{1}{N}\sum_{i = 1}^{N}{\frac{1}{m}\sum_{\tau=1}^{m}{Y^{\alpha}(\tau)}}
\end{equation}
\vspace{0.01in}
\section{YawLoss}

\subsection{Off-Yaw Metric as a Loss Function}
In this section, we show that the Off-Yaw Metric in Eq. (\ref{eq:offyaw}) is differentiable, and is therefore suitable as an auxiliary loss function which penalizes vehicle trajectories that move against the flow of traffic, which we name \textit{YawLoss}. 

We begin with (\ref{eq:offyaw}) and differentiate with respect to network output set of trajectories $T = \{\tau_1, \tau_2,...,\tau_m\}$. For brevity, we abbreviate $x^{\tau}_i, y^{\tau}_i, x^{\tau}_{i+1}, y^{\tau}_{i+1}$ as $\vec{x}^{\tau}_i$. 

\begin{align}
    \nabla \bar{Y}^{\alpha}(T) &= \frac{1}{m} \sum_{\tau=1}^{m}\nabla{Y^{\alpha}(\tau)} \\ 
    &= \frac{1}{mn}\sum_{\tau=1}^{m}\sum_{i=1}^{n}\nabla I(x^\tau_i, y^\tau_i) \delta^{\alpha}(\hat{x}^\tau_i, \hat{y}^\tau_i, \theta(\vec{x}^{\tau}_i))
\end{align}

Since the sum of differentiable functions is differentiable, we continue our analysis with the sum term:
\begin{equation}
g(\vec{x}_i^{\tau}) = \nabla I(\hat{x}^\tau_i, \hat{y}^\tau_i) \delta^{\alpha}(\hat{x}^\tau_i, \hat{y}^\tau_i, \theta(\vec{x}^{\tau}_i))
\end{equation}

Computing the gradient, first for $x_i^{\tau}$, we find:
\begin{align}
    \nonumber
    \frac{\partial{g}}{\partial{x_i^{\tau}}} = &\frac{\partial{I(\hat{x}^\tau_i, \hat{y}^\tau_i)}}{\partial{x_i^{\tau}}}\delta^{\alpha}(\hat{x}^\tau_i, \hat{y}^\tau_i, \theta(\vec{x}^{\tau}_i)) \\ 
    &+ I(\hat{x}^\tau_i, \hat{y}^\tau_i)\frac{\partial{\delta^{\alpha}(\hat{x}^\tau_i, \hat{y}^\tau_i, \theta(\vec{x}^{\tau}_i))}}{\partial{x_i^{\tau}}}
\end{align}
Because the value of the function $I$ in the expression 

\begin{equation}
    \frac{\partial{I(\hat{x}^\tau_i, \hat{y}^\tau_i)}}{\partial{x_i^{\tau}}}
\end{equation}
\noindent can only take on values of 0 or 1, the gradient function is simply 0 when the vehicle remains on-road or off-road, and the positive or negative reciprocal of the displacement of $x^{\tau}_i$ otherwise; in any case, defined for all input. 

The function $\delta^{\alpha}$ of   
\begin{equation}
    \frac{\partial{\delta^{\alpha}(\hat{x}^\tau_i, \hat{y}^\tau_i, \theta(\vec{x}^{\tau}_i))}}{\partial{x_i^{\tau}}} 
\end{equation}

 \noindent will always give a value in the range [0, 360), so the rate change relative to any distance that the $x_i^{\tau}$ coordinate is displaced will be defined for all input. 
The same cases can be extended to the remaining three variables of differentiation ($y_i, x_{i+1}, y_{i+1}$), thus making the function $\bar{Y}^{\alpha}(T)$ differentiable and therefore a suitable loss function.

\medskip
Ultimately, this auxiliary loss function encourages trajectories to stay near lanes whose headings they align with, and to adjust their own headings to more closely match that of the nearest lane. For each midpoint between points in a trajectory, the loss function's value increases as the yaw associated with the midpoint turns further from the yaw of the nearest lane, reaching a maximum when this difference is $180^\circ$, and a minimum at $0^\circ$ or within the provided tolerance threshold. 

Because a map-based trajectory should (in regular cases) not predict movement against the flow of traffic, the loss function is appropriate to apply to all trajectories in multimodal models such as MTP and Multipath. This is a unique quality, as other loss functions may be used only for the most-likely mode per sample, to prevent changing a model's prediction for non-relevant trajectories. For example, a trained MTP model may produce a spread of trajectories covering many possible actions as an intersection (left turn, straight, right turn, U-turn, etc.), but during training, MTP loss would rightfully make adjustments to only its left-turn modes when examining a left-turn sample. By contrast, YawLoss enforces a real-world constraint which must apply across all trajectories (that is, a car must not turn into oncoming traffic), and is therefore applicable to every mode simultaneously. 

\section{Experimental Analysis and Evaluations}

\subsection{Dataset}

To train and evaluate our model, we use the public nuScenes dataset \cite{nuscenes2019}, containing real-world inner-city drives conducted in Boston and Singapore, where each sample includes a raster of the surrounding map, vehicle state information (velocity, acceleration, heading), and target trajectory. Ego vehicle information is encoded with a color index (in this case, red) while surround vehicles are provided a different color (yellow); darker shade renderings of the respective vehicle are used to indicate vehicle location at past time samples, as a means of illustrating prior motion from a single image. Our data is divided using the official benchmark split for the nuScenes prediction challenge; in total, we used 29889 instances in the train set, 7905 instances in the validation set, and 8397 instances in the test set.

\subsection{Network Architecture and Implementation Details}

As introduced in Section II, we perform experiments using both the MTP network defined in \cite{cui2019multimodal} and the MultiPath network defined in \cite{IEEEexample:multipath}. For our experiments, we use a network output of 15 modes with 12 predicted points per mode (representing 6 seconds of travel) for MTP, and 12 predicted offsets per anchor for MultiPath. We use a base CNN of ResNet-50 \cite{DBLP:journals/corr/HeZRS15}. In accordance with the expected input to ResNet with ImageNet dataset pretraining, we normalize our rasterized map images in RGB space prior to training. 
We use the classification and regression loss functions as defined in \cite{cui2019multimodal}, with an additive term for the lane heading auxiliary loss (YawLoss) defined in this work, with a scaling hyperparameter of 1. 

With earlier described rasterized map physical dimensions of 50 meters x 50 meters, using a scale of 0.1 meters per pixel, we assume the lane and trajectory to be approximately straight (i.e. of single uniform heading) on the pixel scale. Each scene map contains information on lane placement and heading, drivable area, and surrounding vehicles and pedestrians. Vehicle state is provided as a three-dimensional input. We use a batch size of 16 and Adam optimizer \cite{DBLP:journals/corr/KingmaB14}, implemented using PyTorch \cite{baydin2017automatic}.

\subsection{Reducing Network Training Time \& Memory Requirements with Secondary Maps}

Calculating this loss per-sample can be computationally expensive. For every predicted mode of each sample instance, it is required to find the $L2$-nearest lane point to each midpoint on the predicted trajectory, with predictions changing on every iteration. 

This computational hurdle can be lowered through preprocessing; for each instance map, which in our case  extends 10 meters behind the vehicle, 40 meters ahead, 25 meters left, and 25 meters right, we generate a secondary orientation map, covering a larger area to account for trajectories which leave the original map. This secondary map extends 20 meters behind the vehicle, 80 meters ahead, 50 meters left, and 50 meters right. On this map, each pixel location is assigned a value which equals the orientation of the nearest lane point. 

These secondary maps are generated and saved for each data sample prior to training. Each grid location on the map represents a heading from the continuous range [0, 360) degrees in the global frame. To represent each grid location as a 64-bit floating point value can quickly become storage intensive for a large set of 500x500 maps. However, only a coarse precision of the angle is required for this problem; we would never consider a driver to be going the `wrong way' if their heading was off by just a few degrees. For this reason, a representation with precision only to the scale of degrees is appropriate for this problem. With this in mind, we can create a data-efficient representation which encodes each heading as an 8-bit grayscale integer pixel value in the range [1, 255], with the value of 0 reserved for map locations corresponding to intersections. Headings are mapped from range [0, 360) degree values to [1, 255] grayscale values as follows:

\begin{equation}
    \theta_{map} = 1 + \lfloor \frac{254}{360} \theta \rceil.
\end{equation}

\begin{figure}
    \centering
    \includegraphics[scale=0.33]{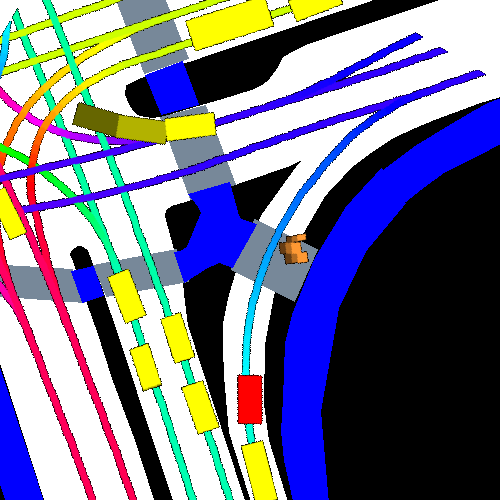} 
    \includegraphics[scale=.66]{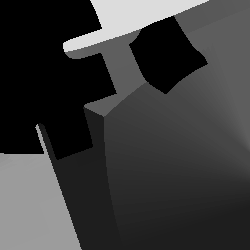}
    \caption{Left: The rasterized bird's-eye-view RGB input map for a sample. 
    Right: The secondary map for the same sample, where each pixel maps to the approximate heading of the nearest lane, or zero if in an intersection. Each pixel's shade of gray represents the orientation of the nearest lane to the pixel. Areas of intersection (i.e. multiple lanes converging or crossing) are given a value of 0 in the grayscale map to represent the ambiguity between the nearest lane and the driver's intended lane in such situations.}
    \label{fig:my_label}
\end{figure}

Using the above function, we assign to each point on the secondary map the mapped value of the heading of the $L2$-nearest lane, illustrated in Fig. \ref{fig:my_label}. During training, when inversely mapping from grayscale integer to degrees, there is a loss of precision that occurs as the 360 degrees are mapped to 254 values. In this sense, each `bin' of the data representation actually represents a span of approximately $1.417^{\circ}$, a reasonable precision for this task.

\begin{figure*}[htp]
\centering
\captionsetup{justification=centering}

\begin{subfigure}[b]{40mm}
\caption*{Ground-truth trajectory from the nuScenes Boston dataset}
\end{subfigure}
\hfill
\begin{subfigure}[b]{40mm}
\caption*{Predicted trajectories using MTP with no auxiliary losses}
\end{subfigure}
\hfill
\begin{subfigure}[b]{40mm}
\caption*{Predicted trajectories using MTP with off-road loss}
\end{subfigure}
\hfill
\begin{subfigure}[b]{40mm}
\caption*{Predicted trajectories using MTP with YawLoss}
\end{subfigure}

\begin{subfigure}[b]{40mm}
\includegraphics[width=40mm]{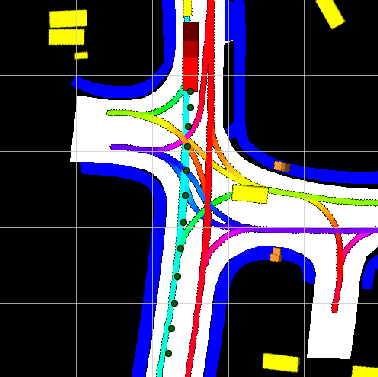}
\caption{}
\end{subfigure}
\hfill
\begin{subfigure}[b]{40mm}
\includegraphics[width=40mm]{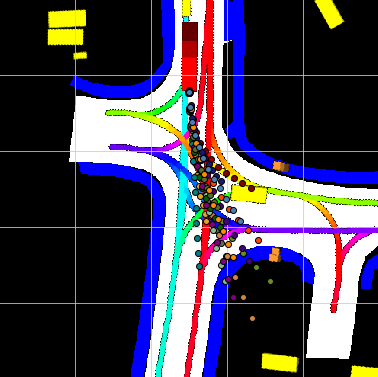}
\caption{}
\end{subfigure}
\hfill
\begin{subfigure}[b]{40mm}
\includegraphics[width=40mm]{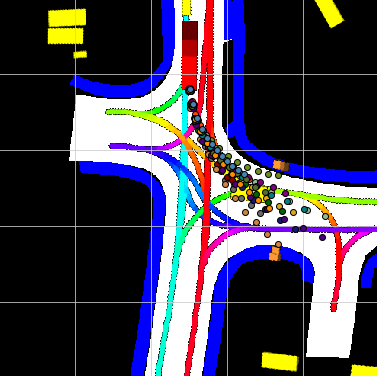}
\caption{}
\end{subfigure}
\hfill
\begin{subfigure}[b]{40mm}
\includegraphics[width=40mm]{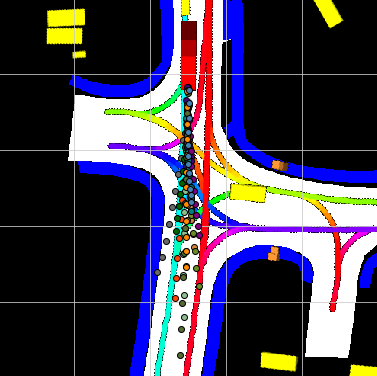}
\caption{}
\end{subfigure}

\begin{subfigure}[b]{40mm}
\includegraphics[width=40mm]{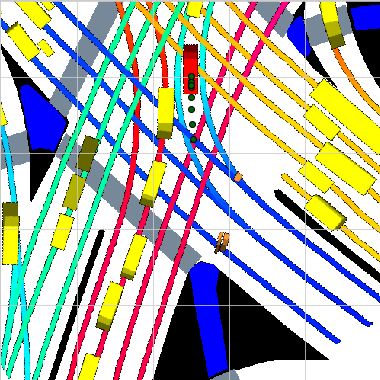}
\caption{}
\end{subfigure}
\hfill
\begin{subfigure}[b]{40mm}
\includegraphics[width=48mm]{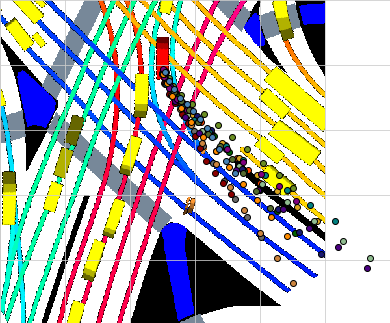}
\caption{}
\end{subfigure}
\hfill
\begin{subfigure}[b]{40mm}
\includegraphics[width=40mm]{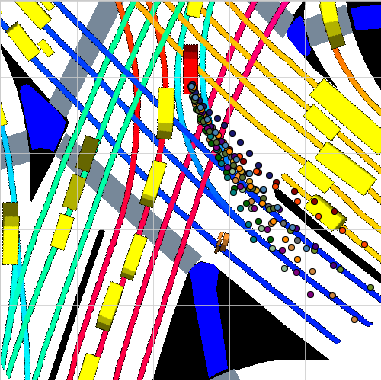}
\caption{}
\end{subfigure}
\hfill
\begin{subfigure}[b]{40mm}
\includegraphics[width=40mm]{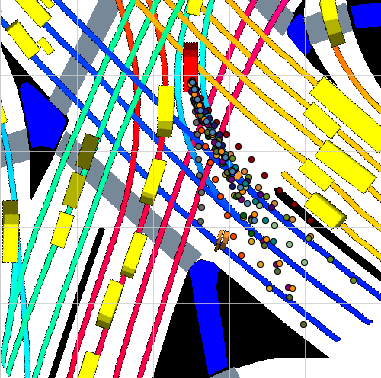}
\caption{}
\end{subfigure}

\begin{subfigure}[b]{40mm}
\includegraphics[width=54mm]{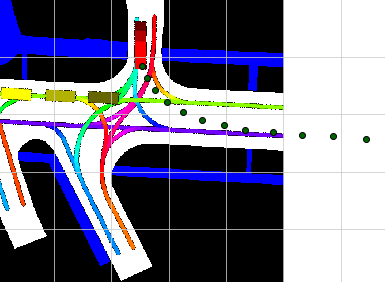}
\caption{}
\end{subfigure}
\hfill
\begin{subfigure}[b]{40mm}
\includegraphics[width=40mm]{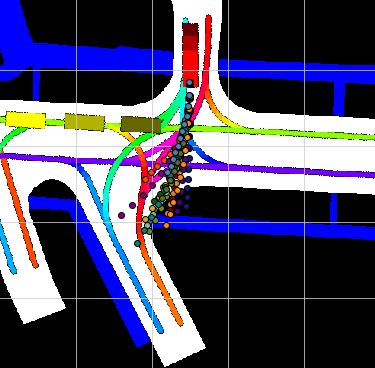}
\caption{}
\end{subfigure}
\hfill
\begin{subfigure}[b]{40mm}
\includegraphics[width=40mm]{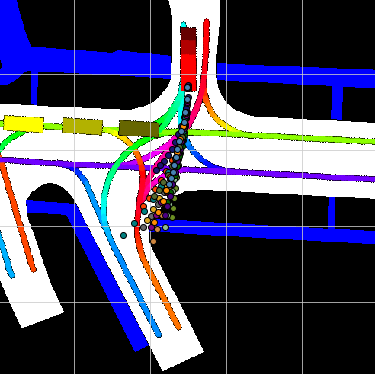}
\caption{}
\end{subfigure}
\hfill
\begin{subfigure}[b]{40mm}
\includegraphics[width=40mm]{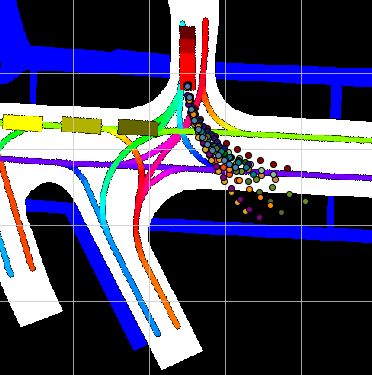}
\caption{}
\end{subfigure}
\captionsetup{justification=default}
\caption{Three examples of improved trajectory prediction using YawLoss. Each row represents a naturalistic Boston driving scenario from the nuScenes dataset. The first column contains the ground-truth trajectory, and the second column contains predictions by the standard MTP model. In the third column, the model is extended with off-road loss. While all three off-road loss examples show trajectories closer to a drivable area, trajectories in the third column are incorrectly pushed into oncoming traffic. By contrast, examples trained with YawLoss (fourth column) show trajectories restored to the drivable area into lanes with the correct heading.}
\label{fig:comparison1}
\end{figure*}

\begin{table*}[t]
\caption{Results of comparative analysis of different models on the nuScenes dataset, over a prediction horizon of 6-seconds. Variants of MultiPath and MTP are grouped for comparison on nine selected metrics. In general, models using YawLoss (this research) improve over the baseline on most metrics.}
\begin{adjustbox}{width=\textwidth}
\begin{tabular}{c c c c c c c c c c c c} 
 \Xhline{3\arrayrulewidth}
 & $MinADE_1 \downarrow $  & $MinADE_5 \downarrow $ & $MinADE_{10} \downarrow $ & $MinFDE_1 \downarrow $ & $MinFDE_5 \downarrow $ & $MinFDE_{10} \downarrow $ & $Miss Rate_{5,2} \downarrow $ & $Miss Rate_{10,2} \downarrow $ & $ Off-Road Rate \downarrow $ \\ 
\Xhline{3\arrayrulewidth}
 Constant Velocity, Yaw & 4.61 & 4.61 & 4.61 & 11.21 & 11.21 & 11.21 & 0.91 & 0.91 & 0.14  \\ 
 \hline
 Physics Oracle & 3.69 & 3.69 & 3.69 & 9.06 & 9.06 & 9.06 & 0.88 & 0.88 & 0.12  \\
 \Xhline{3\arrayrulewidth}
 MultiPath & 4.06 & \hl{\textbf{1.63}} & \hl{\textbf{1.50}} & 9.34 & 3.36 & 3.00 & \hl{\textbf{0.75}} & \hl{\textbf{0.74}} & 0.40  \\
 \hline
 MultiPath with YawLoss & \hl{\textbf{3.95}} & \hl{\textbf{1.63}} & \hl{\textbf{1.50}} & \hl{\textbf{9.08}} & \hl{\textbf{3.33}} & \hl{\textbf{2.95}} & \hl{\textbf{0.75}} & \hl{\textbf{0.74}} & \hl{\textbf{0.38}}  \\
 \Xhline{3\arrayrulewidth}

 MTP & 4.59 & 2.44 & \hl{\textbf{1.57}} & 10.75 & 5.37 & 3.16 & 0.70 & \hl{\textbf{0.55}} & 0.11 \\
 \hline
 MTP with Off Road Loss & 4.51 & \hl{\textbf{2.16}} & 1.60 & 10.44 & \hl{\textbf{4.73}} & 3.23 & 0.72 & 0.58 & 0.13  \\
 \hline
 
 MTP with YawLoss & \hl{\textbf{4.16}} & 2.23 & \hl{\textbf{1.57}} & \hl{\textbf{9.65}} & 4.85 & \hl{\textbf{3.14}} & \hl{\textbf{0.69}} & 0.56 & \hl{\textbf{0.10}} \\ 
\Xhline{3\arrayrulewidth}
\end{tabular}
\end{adjustbox}
\label{tab:results}
\end{table*}

\begin{table}[t]
\caption{Results of comparative analysis of off-yaw rate between two versions of the nuScenes dataset, over a prediction horizon of 6-seconds. The first version is the full validation set, and the second version excludes trajectories whose ground truth contains points within an intersection or off the rasterized map.}
\begin{center}
\begin{tabular}{c c c} 
 \Xhline{2\arrayrulewidth}
 Off-Yaw Rates [rad] $\downarrow$: & All Scenarios & No Intersections \\ 
 \Xhline{2\arrayrulewidth}
 MultiPath & 0.375 & 0.280 \\
 \hline
 MultiPath with YawLoss & \textbf{0.367} & \textbf{0.276} \\
\Xhline{2\arrayrulewidth}
 MTP & \textbf{0.114} & 0.110 \\
 \hline
 MTP with YawLoss & 0.124 & \textbf{0.097} \\ 
 \Xhline{2\arrayrulewidth}
\end{tabular}
\end{center}
\label{tab:results2}
\end{table}

\begin{figure}[htp]
\centering
\includegraphics[width=35mm]{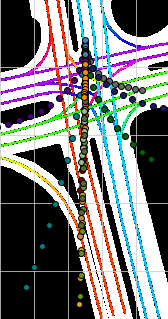} 
\includegraphics[width=34.5mm]{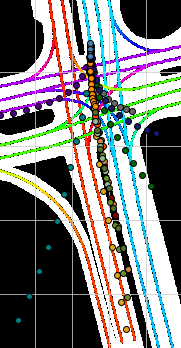} 
\caption{Predicted trajectories using MultiPath with (left) and without (right) YawLoss, illustrating the influence of intersection and off-map points on the calculation of the Off-Yaw Metric. The trajectories in the left image have 31 more intersection points (which contribute no penalty to the metric), so the left trajectories have a much lower off-yaw rate (0.26 difference) despite being less aligned to their lanes.}
\label{fig:comparison2}
\end{figure}

\subsection{Baselines and Metrics}

Results are shown in comparison to the following baselines:
\begin{itemize}
    \item Constant Velocity, Yaw: The predicted trajectory is a continuation of the vehicle's current velocity and heading. 
    \item Physics Oracle: As introduced in \cite{phan2020covernet}, the proposed trajectory is selected as the best trajectory from four dynamics models: constant velocity and yaw, constant velocity and yaw rate, constant acceleration and yaw, and constant acceleration and yaw rate. Note that this method is not used to make predictions, but rather provides a reference benchmark to four simple physical models, to illustrate improvement from models which account for more complex maneuvers.
    \item MultiPath: The predicted trajectories are the output of the MultiPath model, as described in \cite{IEEEexample:multipath}.
    \item MTP: The predicted trajectories are the output of the original MTP model, as described in \cite{cui2019multimodal}.
\end{itemize}

Reported metrics include minimum average displacement error ($MinADE_k$), minimum final displacement error ($MinFDE_k$), miss rate at 2 meters ($Miss Rate_{k,2}$), off-road rate, and off-yaw rate (the new metric as defined in this paper, measuring the amount of positive angular difference of predicted trajectories from the nearest lane yaw, averaged over all agents. $MinADE_k$,  $MinFDE_k$, and $Miss Rate_{k,2}$ are taken over the $k$ most probable trajectories, for $k = 1, 5, \text{and } 10$. While $k = 1$ is generally helpful to evaluate precision of trajectory prediction, in cases when the most probable trajectory is incorrect, the metric value for trajectories comprised of modes which take the average of multiple paths (e.g. going straight when deciding between a left and a right turn) will outperform an incorrect turn for $k = 1$. Thus, including higher $k$ values evaluates whether the model has developed diversity of modes. In all cases, optimal values minimize these displacement errors. 

\subsection{Quantitative Results}
We compare our extension of the MTP and MultiPath models to the various baselines in Table \ref{tab:results}. Our MTP model outperforms or matches the non-extended MTP model on 8 of the 9 reported metrics, the exception being a .01 increase of Miss Rate at 2 m for $k = 10$. In contrast, the MTP model with off-road loss outperforms the baseline on just 4 of the 9 reported metrics. Our MultiPath model outperforms or matches the non-extended MultiPath model on all 9 reported metrics. These improvements suggest that using YawLoss to extend the models created trajectories which have points more closely aligned to the ground truth trajectories and better maintain paths on drivable regions. Additionally, the predicted final location of the vehicle is more close to the known destination.

Qualitative illustrations comparing the effects of Off-Road Loss and YawLoss on an MTP base model are shown in Fig. \ref{fig:comparison1}. As the scenes demonstrate, while off-road loss is effective at bringing trajectories closer to the drivable area, YawLoss is more effective at bringing trajectories closer to the drivable area with the correct heading. 

It is interesting to note that off-yaw rates are similar regardless of auxiliary loss, and in fact sometimes slightly higher when using YawLoss. By Equation \ref{eq:indic}, a non-linearity is introduced for points within an intersection or outside the map region, where the additive rate term is dropped to 0 instantaneously. Thus, it is possible that trajectories at higher velocity (i.e. more likely to leave the drivable region) and trajectories comprising of intersection points, even if further from the ground truth, may receive a lower off-yaw measure than an correct trajectory which leaves the intersection or stays within the map. An illustration of this behavior is shown in Fig. \ref{fig:comparison2}, with a qualitative comparison of the complete dataset with and without intersection and off-map points provided in Table \ref{tab:results2}. For this dataset, the off-yaw rate rises when using YawLoss, while it expectedly decreases when we only consider samples that do not contain this sudden non-linearity. Thus, as a comparative tool, YawLoss is most useful when comparing samples with the same number of non-intersection, on-map points. 

\section{Concluding Remarks}

In this paper we presented an auxiliary loss function which may be used to augment the performance of existing models for vehicle trajectory prediction in urban environments. This lane heading loss function leverages the expectation that vehicles follow the direction ascribed to roadway lanes at all times, with exception for corrective maneuvers, lane changes, and intersection crossings. This loss function applies to all predicted modes, since no mode should predict driving opposite the lane direction. Experiments showed that extending the benchmark MTP model with the lane heading auxiliary loss outperforms the model's original classification and regression losses. 

A possible extension of this work would be the application of the lane heading auxiliary loss to other existing deep learning models, in tandem with other auxiliary losses such as off-road loss. Another possibility for future investigation is the tuning of the angular difference threshold and weighting using agent dynamics and scene context. Finally, in our future work, we intend to design a methodology for quantifying nearest lane heading within an intersection or outside of the drivable area to reduce the effect of this non-linearity on training and metric reporting. 

As stated by Daily et al. \cite{daily2017self}, ``Self-driving and highly automated vehicles must navigate smoothly and avoid obstacles, while accurately understanding the highly complex semantic interpretation of scene and dynamic activities." While convolutional neural networks and other data-driven approaches may be effective at repeating known patterns, there is a lost element of explainability which is crucial towards public safety and adoption. By encoding familiar driving expectations through the introduced off-yaw rate metric and YawLoss, we initiate a step towards autonomous vehicle computational models which can both learn and explain.

\section*{Acknowledgements}
We are grateful to our reviewers for their valuable comments. We also thank LISA sponsors and colleagues for their support, with special thanks to Kaouther Messaoud for assistance in baseline implementation. 

\bibliographystyle{./bibliography/IEEEtran}
\bibliography{./bibliography/IEEEexample}

\begin{thebibliography}{10}
\providecommand{\url}[1]{#1}
\csname url@samestyle\endcsname
\providecommand{\newblock}{\relax}
\providecommand{\bibinfo}[2]{#2}
\providecommand{\BIBentrySTDinterwordspacing}{\spaceskip=0pt\relax}
\providecommand{\BIBentryALTinterwordstretchfactor}{4}
\providecommand{\BIBentryALTinterwordspacing}{\spaceskip=\fontdimen2\font plus
\BIBentryALTinterwordstretchfactor\fontdimen3\font minus
  \fontdimen4\font\relax}
\providecommand{\BIBforeignlanguage}[2]{{%
\expandafter\ifx\csname l@#1\endcsname\relax
\typeout{** WARNING: IEEEtran.bst: No hyphenation pattern has been}%
\typeout{** loaded for the language `#1'. Using the pattern for}%
\typeout{** the default language instead.}%
\else
\language=\csname l@#1\endcsname
\fi
#2}}
\providecommand{\BIBdecl}{\relax}
\BIBdecl

\bibitem{deo2018multi}
N.~Deo and M.~M. Trivedi, ``Multi-modal trajectory prediction of surrounding
  vehicles with maneuver based lstms,'' in \emph{2018 IEEE Intelligent Vehicles
  Symposium (IV)}.\hskip 1em plus 0.5em minus 0.4em\relax IEEE, 2018, pp.
  1179--1184.

\bibitem{deo2018convolutional}
------, ``Convolutional social pooling for vehicle trajectory prediction,'' in
  \emph{Proceedings of the IEEE Conference on Computer Vision and Pattern
  Recognition Workshops}, 2018, pp. 1468--1476.

\bibitem{cui2019multimodal}
H.~Cui, V.~Radosavljevic, F.-C. Chou, T.-H. Lin, T.~Nguyen, T.-K. Huang,
  J.~Schneider, and N.~Djuric, ``Multimodal trajectory predictions for
  autonomous driving using deep convolutional networks,'' in \emph{2019
  International Conference on Robotics and Automation (ICRA)}.\hskip 1em plus
  0.5em minus 0.4em\relax IEEE, 2019, pp. 2090--2096.

\bibitem{IEEEexample:multipath}
Y.~Chai, B.~Sapp, M.~Bansal, and D.~Anguelov, ``Multipath: Multiple
  probabilistic anchor trajectory hypotheses for behavior prediction,''
  \emph{arXiv preprint arXiv:1910.05449}, 2019.

\bibitem{ridel2020scene}
D.~Ridel, N.~Deo, D.~Wolf, and M.~Trivedi, ``Scene compliant trajectory
  forecast with agent-centric spatio-temporal grids,'' \emph{IEEE Robotics and
  Automation Letters}, vol.~5, no.~2, pp. 2816--2823, 2020.

\bibitem{messaoud2020multi}
K.~Messaoud, N.~Deo, M.~M. Trivedi, and F.~Nashashibi, ``Multi-head attention
  with joint agent-map representation for trajectory prediction in autonomous
  driving,'' \emph{arXiv preprint arXiv:2005.02545}, 2020.

\bibitem{casas2020importance}
S.~Casas, C.~Gulino, S.~Suo, and R.~Urtasun, ``The importance of prior
  knowledge in precise multimodal prediction,'' \emph{arXiv preprint
  arXiv:2006.02636}, 2020.

\bibitem{liang2020learning}
M.~Liang, B.~Yang, R.~Hu, Y.~Chen, R.~Liao, S.~Feng, and R.~Urtasun, ``Learning
  lane graph representations for motion forecasting,'' \emph{arXiv preprint
  arXiv:2007.13732}, 2020.

\bibitem{gupta2018social}
A.~Gupta, J.~Johnson, L.~Fei-Fei, S.~Savarese, and A.~Alahi, ``Social gan:
  Socially acceptable trajectories with generative adversarial networks,'' in
  \emph{Proceedings of the IEEE Conference on Computer Vision and Pattern
  Recognition}, 2018, pp. 2255--2264.

\bibitem{sadeghian2019sophie}
A.~Sadeghian, V.~Kosaraju, A.~Sadeghian, N.~Hirose, H.~Rezatofighi, and
  S.~Savarese, ``Sophie: An attentive gan for predicting paths compliant to
  social and physical constraints,'' in \emph{Proceedings of the IEEE
  Conference on Computer Vision and Pattern Recognition}, 2019, pp. 1349--1358.

\bibitem{amirian2019social}
J.~Amirian, J.-B. Hayet, and J.~Pettr{\'e}, ``Social ways: Learning multi-modal
  distributions of pedestrian trajectories with gans,'' in \emph{Proceedings of
  the IEEE Conference on Computer Vision and Pattern Recognition Workshops},
  2019, pp. 0--0.

\bibitem{kosaraju2019social}
V.~Kosaraju, A.~Sadeghian, R.~Mart{\'\i}n-Mart{\'\i}n, I.~Reid, H.~Rezatofighi,
  and S.~Savarese, ``Social-bigat: Multimodal trajectory forecasting using
  bicycle-gan and graph attention networks,'' in \emph{Advances in Neural
  Information Processing Systems}, 2019, pp. 137--146.

\bibitem{zhao2019multi}
T.~Zhao, Y.~Xu, M.~Monfort, W.~Choi, C.~Baker, Y.~Zhao, Y.~Wang, and Y.~N. Wu,
  ``Multi-agent tensor fusion for contextual trajectory prediction,'' in
  \emph{Proceedings of the IEEE Conference on Computer Vision and Pattern
  Recognition}, 2019, pp. 12\,126--12\,134.

\bibitem{wang2020improving}
E.~Wang, H.~Cui, S.~Yalamanchi, M.~Moorthy, F.-C. Chou, and N.~Djuric,
  ``Improving movement predictions of traffic actors in bird's-eye view models
  using gans and differentiable trajectory rasterization,'' \emph{arXiv
  preprint arXiv:2004.06247}, 2020.

\bibitem{lee2017desire}
N.~Lee, W.~Choi, P.~Vernaza, C.~B. Choy, P.~H. Torr, and M.~Chandraker,
  ``Desire: Distant future prediction in dynamic scenes with interacting
  agents,'' in \emph{Proceedings of the IEEE Conference on Computer Vision and
  Pattern Recognition}, 2017, pp. 336--345.

\bibitem{ivanovic2019trajectron}
B.~Ivanovic and M.~Pavone, ``The trajectron: Probabilistic multi-agent
  trajectory modeling with dynamic spatiotemporal graphs,'' in
  \emph{Proceedings of the IEEE International Conference on Computer Vision},
  2019, pp. 2375--2384.

\bibitem{salzmann2020trajectron++}
T.~Salzmann, B.~Ivanovic, P.~Chakravarty, and M.~Pavone, ``Trajectron++:
  Multi-agent generative trajectory forecasting with heterogeneous data for
  control,'' \emph{arXiv preprint arXiv:2001.03093}, 2020.

\bibitem{casas2020implicit}
S.~Casas, C.~Gulino, S.~Suo, K.~Luo, R.~Liao, and R.~Urtasun, ``Implicit latent
  variable model for scene-consistent motion forecasting,'' \emph{arXiv
  preprint arXiv:2007.12036}, 2020.

\bibitem{rhinehart2018r2p2}
N.~Rhinehart, K.~M. Kitani, and P.~Vernaza, ``R2p2: A reparameterized
  pushforward policy for diverse, precise generative path forecasting,'' in
  \emph{Proceedings of the European Conference on Computer Vision (ECCV)},
  2018, pp. 772--788.

\bibitem{rhinehart2019precog}
N.~Rhinehart, R.~McAllister, K.~Kitani, and S.~Levine, ``Precog: Prediction
  conditioned on goals in visual multi-agent settings,'' in \emph{Proceedings
  of the IEEE International Conference on Computer Vision}, 2019, pp.
  2821--2830.

\bibitem{bhattacharyya2019conditional}
A.~Bhattacharyya, M.~Hanselmann, M.~Fritz, B.~Schiele, and C.-N. Straehle,
  ``Conditional flow variational autoencoders for structured sequence
  prediction,'' \emph{arXiv preprint arXiv:1908.09008}, 2019.

\bibitem{bhattacharyya2020haar}
A.~Bhattacharyya, C.-N. Straehle, M.~Fritz, and B.~Schiele, ``Haar wavelet
  based block autoregressive flows for trajectories,'' \emph{arXiv preprint
  arXiv:2009.09878}, 2020.

\bibitem{ziebart2009planning}
B.~D. Ziebart, N.~Ratliff, G.~Gallagher, C.~Mertz, K.~Peterson, J.~A. Bagnell,
  M.~Hebert, A.~K. Dey, and S.~Srinivasa, ``Planning-based prediction for
  pedestrians,'' in \emph{2009 IEEE/RSJ International Conference on Intelligent
  Robots and Systems}.\hskip 1em plus 0.5em minus 0.4em\relax IEEE, 2009, pp.
  3931--3936.

\bibitem{kitani2012activity}
K.~M. Kitani, B.~D. Ziebart, J.~A. Bagnell, and M.~Hebert, ``Activity
  forecasting,'' in \emph{European Conference on Computer Vision}.\hskip 1em
  plus 0.5em minus 0.4em\relax Springer, 2012, pp. 201--214.

\bibitem{zhang2018integrating}
Y.~Zhang, W.~Wang, R.~Bonatti, D.~Maturana, and S.~Scherer, ``Integrating
  kinematics and environment context into deep inverse reinforcement learning
  for predicting off-road vehicle trajectories,'' \emph{arXiv preprint
  arXiv:1810.07225}, 2018.

\bibitem{deo2020trajectory}
N.~Deo and M.~M. Trivedi, ``Trajectory forecasts in unknown environments
  conditioned on grid-based plans,'' \emph{arXiv preprint arXiv:2001.00735},
  2020.

\bibitem{niedoba2019improving}
M.~Niedoba, H.~Cui, K.~Luo, D.~Hegde, F.-C. Chou, and N.~Djuric, ``Improving
  movement prediction of traffic actors using off-road loss and bias
  mitigation,'' in \emph{Workshop on'Machine Learning for Autonomous Driving'at
  Conference on Neural Information Processing Systems}, 2019.

\bibitem{boulton2020motion}
F.~A. Boulton, E.~C. Grigore, and E.~M. Wolff, ``Motion prediction using
  trajectory sets and self-driving domain knowledge,'' \emph{arXiv preprint
  arXiv:2006.04767}, 2020.

\bibitem{rudenko2020human}
A.~Rudenko, L.~Palmieri, M.~Herman, K.~M. Kitani, D.~M. Gavrila, and K.~O.
  Arras, ``Human motion trajectory prediction: A survey,'' \emph{The
  International Journal of Robotics Research}, vol.~39, no.~8, pp. 895--935,
  2020.

\bibitem{ridel2018literature}
D.~Ridel, E.~Rehder, M.~Lauer, C.~Stiller, and D.~Wolf, ``A literature review
  on the prediction of pedestrian behavior in urban scenarios,'' in \emph{2018
  21st International Conference on Intelligent Transportation Systems
  (ITSC)}.\hskip 1em plus 0.5em minus 0.4em\relax IEEE, 2018, pp. 3105--3112.

\bibitem{phan2020covernet}
T.~Phan-Minh, E.~C. Grigore, F.~A. Boulton, O.~Beijbom, and E.~M. Wolff,
  ``Covernet: Multimodal behavior prediction using trajectory sets,'' in
  \emph{Proceedings of the IEEE/CVF Conference on Computer Vision and Pattern
  Recognition}, 2020, pp. 14\,074--14\,083.

\bibitem{casas2020spagnn}
S.~Casas, C.~Gulino, R.~Liao, and R.~Urtasun, ``Spagnn: Spatially-aware graph
  neural networks for relational behavior forecasting from sensor data,'' in
  \emph{2020 IEEE International Conference on Robotics and Automation
  (ICRA)}.\hskip 1em plus 0.5em minus 0.4em\relax IEEE, 2020, pp. 9491--9497.

\bibitem{cui2020deep}
H.~Cui, T.~Nguyen, F.-C. Chou, T.-H. Lin, J.~Schneider, D.~Bradley, and
  N.~Djuric, ``Deep kinematic models for kinematically feasible vehicle
  trajectory predictions,'' in \emph{International Conference on Robotics and
  Automation (ICRA)}.\hskip 1em plus 0.5em minus 0.4em\relax IEEE, 2020, pp.
  10\,563--10\,569.

\bibitem{chang2019argoverse}
M.-F. Chang, J.~Lambert, P.~Sangkloy, J.~Singh, S.~Bak, A.~Hartnett, D.~Wang,
  P.~Carr, S.~Lucey, D.~Ramanan \emph{et~al.}, ``Argoverse: 3d tracking and
  forecasting with rich maps,'' in \emph{Proceedings of the IEEE Conference on
  Computer Vision and Pattern Recognition}, 2019, pp. 8748--8757.

\bibitem{nuscenes2019}
H.~Caesar, V.~Bankiti, A.~H. Lang, S.~Vora, V.~E. Liong, Q.~Xu, A.~Krishnan,
  Y.~Pan, G.~Baldan, and O.~Beijbom, ``nuscenes: A multimodal dataset for
  autonomous driving,'' \emph{arXiv preprint arXiv:1903.11027}, 2019.

\bibitem{DBLP:journals/corr/HeZRS15}
\BIBentryALTinterwordspacing
K.~He, X.~Zhang, S.~Ren, and J.~Sun, ``Deep residual learning for image
  recognition,'' \emph{CoRR}, vol. abs/1512.03385, 2015. [Online]. Available:
  \url{http://arxiv.org/abs/1512.03385}
\BIBentrySTDinterwordspacing

\bibitem{DBLP:journals/corr/KingmaB14}
\BIBentryALTinterwordspacing
D.~P. Kingma and J.~Ba, ``Adam: {A} method for stochastic optimization,'' in
  \emph{3rd International Conference on Learning Representations, {ICLR} 2015,
  San Diego, CA, USA, May 7-9, 2015, Conference Track Proceedings}, Y.~Bengio
  and Y.~LeCun, Eds., 2015. [Online]. Available:
  \url{http://arxiv.org/abs/1412.6980}
\BIBentrySTDinterwordspacing

\bibitem{baydin2017automatic}
A.~G. Baydin, B.~A. Pearlmutter, A.~A. Radul, and J.~M. Siskind, ``Automatic
  differentiation in machine learning: a survey,'' \emph{The Journal of Machine
  Learning Research}, vol.~18, no.~1, pp. 5595--5637, 2017.

\bibitem{daily2017self}
M.~Daily, S.~Medasani, R.~Behringer, and M.~Trivedi, ``Self-driving cars,''
  \emph{Computer}, vol.~50, no.~12, pp. 18--23, 2017.

\end{thebibliography}

\vspace{12pt}

\end{document}